\documentclass[letterpaper, 10 pt, conference]{config/ieeeconf} 
\IEEEoverridecommandlockouts                              % This command is only needed if                               
\makeatletter
\let\NAT@parse\undefined
\makeatother
\usepackage[numbers]{natbib}
\usepackage[bookmarks=true, colorlinks=false,
    linkcolor=black,
    filecolor=mrobota,
    urlcolor=cyan]{hyperref}
\usepackage{amsmath}
\usepackage{amsfonts, amssymb}
\usepackage{algorithm}
\usepackage{soul}
\usepackage{algpseudocode}
\usepackage[dvipsnames]{xcolor}
\usepackage[font=small,skip=3pt]{caption}
\usepackage{subcaption}

\usepackage{graphicx}
\usepackage{csvsimple}
\usepackage{tikz}
\usepackage{pgf}
\usepackage{import}
\usepackage{pgfplots}
\usepackage{pgfplotstable}
\usepgfplotslibrary{fillbetween}
\usepackage{stackengine}
\usepackage{booktabs}
\usepackage{multirow}
\usepackage{siunitx}
\usepackage{svg}
\usepackage{import}

\usepackage{booktabs}
\usepackage{xcolor}
\usepackage{array}
\usepackage{amssymb} % for \checkmark
\usepackage{url}
\pgfplotsset{every x tick label/.append style={font=\tiny, yshift=0.5ex}}
\pgfplotsset{every y tick label/.append style={font=\tiny, xshift=0.5ex}}

\usepackage{booktabs}

\usepackage[utf8]{inputenc}

\DeclareMathOperator*{\argmax}{arg\,max}

\newtheorem{problem}{Problem}
\newtheorem{definition}{Definition}
\usepackage{stackengine}
\usepackage{color}
\usepackage{todonotes}
\usepackage{subcaption}
\usepackage{amssymb}
\usepackage{orcidlink}
\newcommand{\noboxorcid}[1]{%
  {\hypersetup{pdfborder={0 0 0}}\href{https://orcid.org/#1}{\orcidlink{#1}}}%
}
\title{\Large \bf \textcolor{black}{{Multi-Robot Trajectory Planning via Constrained Bayesian Optimization and Local Cost Map Learning with STL-Based Conflict Resolution}}}
 
\author{{Sourav Raxit$^{1}$\noboxorcid{0000-0003-1196-2435}}, {Abdullah Al Redwan Newaz$^{1^*}$\noboxorcid{0000-0003-1140-8119}}, {Jose Fuentes$^{2}$\noboxorcid{0000-0002-6477-5820}}, {Paulo Padrao$^{3}$\noboxorcid{0000-0003-3966-0279}},\\ Ana Cavalcanti$^{2},$ and {Leonardo Bobadilla$^{2}$\noboxorcid{0000-0003-2097-2432}}
\thanks{
This work is supported in part by the U.S. EPA grant BR-02F47801-5010M, NSF grants 2118329, IIS-2024733, IIS-2331908, ONR grant N00014-23-1-2789, DoD grant 78170-RT-REP, the ARL under contract W911NF1920243, and the FDEP grant INV31.}
\thanks{$^{1}$ S. Raxit, and A. A. R. Newaz (* corresponding author) are with the Department of Computer Science, University of New Orleans, New Orleans, LA 70148, USA (email: \{sraxit, aredwann\}@uno.edu).
$^{2}$ J. Fuentes, A. Cavalcanti and L. Bobadilla are with the Knight Foundation School of Computing and Information Sciences, Florida International University, Miami, FL 33199, USA (email:%{\tt\scriptsize \
        \{jfuen099@, acavalca, bobadilla@cs.\}fiu.edu).
$^{3}$ P. Padrao is with  Providence College, Department of Mathematics \& Computer Science, Providence,   RI 02918, USA (email:ppadraol@providence.edu).
        }
}
\begin{document}

\maketitle
\begin{abstract}
%   We address the problem of multi-robot motion planning under Signal Temporal Logic (STL) specifications with kinodynamic constraints. Conventional approaches, such as mixed-integer formulations or model predictive control, often face scalability bottlenecks and limited adaptability in dynamic environments. To overcome these challenges, we propose a two-stage framework that integrates sampling-based planning with formal temporal-logic reasoning. At the single-robot level, we introduce the constrained Bayesian optimization–based tree search (cBOT) planner, which leverages Gaussian process models of local cost maps and feasibility constraints to generate shorter, collision-free trajectories with fewer samples than RRT-based methods. At the multi-robot level, we develop the STL-enhanced Kinodynamic Conflict-Based Search (STL-KCBS) algorithm, which incorporates STL monitoring into conflict detection and resolution, ensuring specification satisfaction while retaining scalability and probabilistic completeness. Benchmarking results demonstrate that the proposed framework improves trajectory efficiency and safety compared to existing methods. Real-world experiments with autonomous surface vehicles further validate its robustness and practical applicability in uncertain environments.
We address multi-robot motion planning under Signal Temporal Logic (STL) specifications with kinodynamic constraints. Exact approaches face scalability bottlenecks and limited adaptability, while conventional sampling-based methods require excessive samples to construct optimal trajectories.
We propose a two-stage framework integrating sampling-based online learning with formal STL reasoning. At the single-robot level, our constrained Bayesian Optimization-based Tree search (cBOT) planner uses a Gaussian process as a surrogate model to learn local cost maps and feasibility constraints, generating shorter collision-free trajectories with fewer samples. At the multi-robot level, our STL-enhanced Kinodynamic Conflict-Based Search (STL-KCBS) algorithm incorporates STL monitoring into conflict detection and resolution, ensuring specification satisfaction while maintaining scalability and probabilistic completeness.
Benchmarking demonstrates improved trajectory efficiency and safety over existing methods. Real-world experiments with autonomous surface vehicles validate robustness and practical applicability in uncertain environments. The STLcBOT Planner will be released as an open-source package, and videos of real-world and simulated experiments are available at \url{https://stlbot.github.io/}.

\end{abstract}
\section{Introduction}
% Temporal logics such as Linear Temporal Logic (LTL)\cite{pnueli1977temporal}, Metric Temporal Logic (MTL)\cite{koymans1990specifying}, and Signal Temporal Logic (STL)\cite{maler2004monitoring} provide expressive frameworks for specifying rich and complex robotic behaviors, making them invaluable for both controller synthesis and formal verification. This expressiveness has motivated the development of numerous motion planning algorithms aimed at satisfying STL specifications\cite{meng2023signal}.

A central challenge in multi-robot systems lies in generating dynamically feasible trajectories that permit robots to simultaneously reach their target regions without colliding with obstacles or with one another. Conventional methods typically rely on explicitly encoding large sets of low-level constraints, leading to brittle solutions with limited adaptability to real-world conditions. 
Signal Temporal Logic (STL)~\cite{maler2004monitoring}, by contrast, offers a powerful declarative formalism that enables high-level specification of temporal and logical requirements, supporting sophisticated behaviors such as International Regulations for Preventing Collisions at
Sea (COLREGs), coordinated motion through intersections, and adaptive responses to environmental changes.

Early STL-based planning approaches frequently employed model predictive control (MPC)\cite{wongpiromsarn2011tulip,sadraddini2015robust,leung2023backpropagation}, which proved effective but computationally expensive when tackling complex formulas and long horizons. To improve scalability, STL constraints have been incorporated into sampling-based planners such as RRT* and its real-time extensions, with robustness metrics guiding the search toward specification-compliant solutions\cite{linard2023real}. Multi-layered frameworks have since integrated kinodynamic constraints and STL through biased exploration and robustness-driven costs, enabling sequential satisfaction of tasks with greater efficiency~\cite{chatrola2025multi}. More recently, reinforcement learning has been combined with STL semantics to address the reward-shaping problem, producing task-aware and robust policies~\cite{kapoor2020model,meng2023signal}. These efforts highlight STL’s potential in single-agent contexts, but extending such capabilities to multi-robot systems introduces substantial new challenges.

% Research in this space initially focused on single-robot planning under temporal logic specifications~\cite{kress2009temporal,wongpiromsarn2011tulip,abbas2019temporal}.

Scaling STL to multi-robot motion planning (MRMP) requires addressing task allocation, inter-agent coordination, and nonlinear or nonholonomic dynamics~\cite{sun2022multi,eappen2024scaling}. Centralized approaches often encode multi-agent STL specifications as mixed-integer linear programs (MILP), allowing implicit task assignment and collision avoidance~\cite{sun2022multi,liu2017communication}. While effective, MILP formulations face scalability bottlenecks inherent to their solvers. Decentralized alternatives have thus emerged: barrier-function-based feedback controllers ensure continuous-time STL satisfaction across agents~\cite{lindemann2020barrier,gundana2021event}, while distributed MPC relaxes infeasible constraints to guarantee recursive feasibility under conflicting tasks~\cite{zhou2022distributed}. Complementing these methods, differentiable STL robustness has enabled graph neural planners to handle heterogeneous teams and cluttered environments with up to 32 robots~\cite{eappen2024scaling}, offering improved scalability over centralized MPC baselines.

Beyond STL-specific work, broader MRMP research provides valuable insights through multi-agent path finding (MAPF) extensions~\cite{sharon2015conflict, ma2019searching, andreychuk2022multi} and sampling-based methods that offer flexibility through collision checking alone. While coupled approaches explore joint state spaces, their synchronized action requirements restrict time optimality, and spatiotemporal variants like Time-Based RRT~\cite{sintov2014time}, Temporal PRM~\cite{huppi2022t}, and ST-RRT*~\cite{grothe2022st} enhance temporal reasoning but struggle in narrow corridors, limiting scalability. Graphs of Convex Sets (GCS) provide a geometry-driven alternative by constructing collision-free convex regions directly, with applications in UAV navigation, non-Euclidean motion planning, and temporal-logic tasks~\cite{marcucci2024fast,cohn2023non,kurtz2023temporal}, while multi-robot extensions show efficiency improvements~\cite{chia2024gcs,marcucci2023motion} despite dynamic coordination challenges that motivate developments like ST-GCS~\cite{tang2025space}. Kinodynamic Conflict-Based Search (K-CBS)~\cite{kottinger2022conflict} decomposes planning into high-level coordination and low-level trajectory generation, achieving probabilistic completeness and scalability but struggling with path optimization due to RRT-based planning and cost map neglect. Though not designed for STL, these methods could be augmented with temporal-logic constraints, bridging discrete conflict resolution and decentralized STL planning. Table~\ref{tab:comparison} summarizes a qualitative comparison of the related methods.

\begin{table}[h!]
\centering
\renewcommand{\arraystretch}{1.2}
\setlength{\tabcolsep}{8pt}
\resizebox{0.48\textwidth}{!}{%
\begin{tabular}{@{}lcccc@{}}
\toprule
\textbf{Method} &
\textbf{Multirobot} &
\textbf{STL Spec} &
\textbf{Learning} &
\textbf{\begin{tabular}[c]{@{}c@{}}Kinodynamic \\ Constraints\end{tabular}} \\ 
\midrule
MILP~\cite{sun2022multi}       & \checkmark & \checkmark & $\times$ & $\times$ \\
STL-RRT*~\cite{grothe2022st,chatrola2025multi}    & $\times$ & \checkmark & $\times$ & $\times$ \\
ST-GCS~\cite{tang2025space}     & \checkmark & \checkmark & $\times$ & $\times$ \\
MPC~\cite{leung2023backpropagation}        & $\times$ & \checkmark & \checkmark & $\times$ \\
GCBF~\cite{eappen2024scaling}       & \checkmark & \checkmark & \checkmark & $\times$ \\
K-CBS~\cite{kottinger2022conflict}      & \checkmark & $\times$ & $\times$ & \checkmark \\
Ours        & \checkmark & \checkmark & \checkmark & \checkmark \\
\bottomrule
\end{tabular}%
}
\caption{Qualitative comparison of the related methods.}
\label{tab:comparison}
\end{table}

\noindent \textbf{Contributions.} 
The main contributions of this paper are as follows: 
($i$) We propose a constrained Bayesian optimization–based tree search (cBOT) algorithm for computing collision-free trajectories for individual robots. The cBOT framework incorporates a local cost map modeled with a Gaussian process (GP) and learns individual constraints using separate GPs. This enables the planner to generate shorter trajectories with fewer samples compared to conventional RRT-based approaches.  
($ii$) We introduce the STL-enhanced Kinodynamic Conflict-Based Search (STL-KCBS) algorithm, which efficiently computes multi-robot trajectories by combining the decoupled, sampling-based cBOT planner with a conflict tree. STL-KCBS inherits the decentralized and scalable properties of the K-CBS framework~\cite{kottinger2022conflict}, while also providing probabilistic completeness guarantees.  
($iii$) We present a comprehensive benchmarking study against existing approaches, demonstrating the advantages of the proposed STL-KCBS planner. A qualitative comparison of related methods is summarized in Table~\ref{tab:comparison}.  
($iv$) We validate the real-world applicability of the STL-KCBS planner through experiments with autonomous surface vehicles operating in a lake environment.  

%\section{Related Work}
%\input{sect/related_work}

% \section{Notation}
% \input{table/notation}

\section{Problem Formulation}
\textcolor{black}{
Consider a team of $ N $ robots operating in a $d$-dimensional workspace 
$W \subset \mathbb{R}^d$, which represents the geometric space of spatial positions}, with the free space defined as 
$ W_{\text{f}} = X_{\text{f}} = X \setminus X_{\text{obs}} $, where 
$ X_{\text{obs}} \subset X \subseteq \mathbb{R}^d $ represents obstacles.

Each robot is initialized at a designated position 
$ \mathbf{x}_i^{\text{init}} \in W $, $ \Pi_i(\mathbf{x}_{i,\text{init}}) = p_i^{\text{init}}$; $\Pi_i:X_i\to W$ is the projection function that maps the robot's state into its position. 
The team’s navigation objective is specified by the multi-agent signal temporal logic (MA-STL) formula $\Psi$ (cf. Definition~\ref{def:mastl}), 
which encodes desired collective behaviors over time. 
Given a real-valued function $\mu: W \to \mathbb{R}$, 
an atomic predicate is defined as $ \pi^\mu $, where a point 
$\mathbf{x} \in W $ satisfies $\mathbf{x} \models \pi^\mu $ 
if $ \mu(\mathbf{x}) \geq 0 $. 

We present notations used throughout: 
$\mathbb{R}$, $\mathbb{R}_{\ge 0}$, and $\mathbb{N}$ denote the sets 
of real, non-negative real, and natural numbers, respectively. 
We write $\mathbb{R}^n $ for the $ n $-dimensional Euclidean space and 
$ \mathbb{R}^{n \times m} $ for the space of real matrices with $ n $ rows 
and $ m $ columns. For $ a \le b $, $ [a,b] $ denotes a time interval, 
and $ t+[a,b] := [t+a, t+b] $. 
% \newcommand{\fun}{\textnormal{\textsc{Fun}}}
% The set $\fun(X,Y)$ denotes all functions from $ X $ to $ Y $.
\textcolor{black}{We denote the set of all functions from $X$ to $Y$ by $Y^{X}$.}

\begin{definition}[Signal Temporal Logic (STL) Semantics]\label{def:stl}
Let $ M \subset \mathbb{R}^d $, 
\textcolor{black}{$ S = M^{\mathbb{R}_{\geq 0}} $} denotes the set of signals, 
and \textcolor{black}{$ F = \mathbb{R}^{\mathbb{R}^{d}} $} the set of real-valued functions from $\mathbb{R}^d$.  
The syntax of STL is given by \cite{sun2022multi}:
\begin{equation*}
\varphi ::= \top \mid \pi^\mu \mid \neg \varphi \mid 
\varphi_1 \wedge \varphi_2 \mid 
\varphi_1 \, \mathcal{U}_{[a,b]} \, \varphi_2,   
\end{equation*}

where  $\top$ is \textsc{True} and $ \mu \in F $.

% The Boolean semantics $ (s, t) \models \varphi $ are defined recursively:

\textcolor{black}{For any signal $s \in S$ and time $t \in \mathbb{R}_{\ge 0}$, the Boolean semantics $(s,t)\models \varphi$ are defined recursively as:}

% (s, t) \models \top &\Leftrightarrow \;\; \textsc{True}, \\
% (s, t) \models \pi^\mu &\Leftrightarrow \mu(s(t)) \ge 0, \\
% (s, t) \models \neg \varphi &\Leftrightarrow \neg \big( (s, t) \models \varphi \big), \\
% (s, t) \models \varphi_1 \wedge \varphi_2 &\Leftrightarrow 
% (s, t) \models \varphi_1 \;\wedge\; (s, t) \models \varphi_2, \\
% (s, t) \models \lozenge_{[a,b]}\varphi &\Leftrightarrow \forall \tau \in t+[a,b], (s,\tau) \models \varphi  \\
% (s, t) \models \square_{[a,b]}\varphi &\Leftrightarrow \exists \tau \in t+[a,b], (s,\tau) \models \varphi\\
% (s,t)\models \varphi_1 \, U_{[a,b]} \, \varphi_2
% \iff
\vspace{-12 pt}
\textcolor{black}{
\[
\begin{aligned}
(s,t)\models \top &\Leftrightarrow \textsc{True}, \\
(s,t)\models \pi^\mu &\Leftrightarrow \mu(s(t)) \ge 0, \\
(s,t)\models \neg \varphi &\Leftrightarrow \neg\big((s,t)\models \varphi\big), \\
(s,t)\models \varphi_1 \wedge \varphi_2 
&\Leftrightarrow \big((s,t)\models \varphi_1\big)\ \wedge\ \big((s,t)\models \varphi_2\big), \\
(s,t)\models \lozenge_{[a,b]}\varphi 
&\Leftrightarrow \exists \tau \in t+[a,b]\ \text{s.t.}\ (s,\tau)\models \varphi, \\
(s,t)\models \square_{[a,b]}\varphi 
&\Leftrightarrow \forall \tau \in t+[a,b],\ (s,\tau)\models \varphi, \\
(s,t)\models \varphi_1\, U_{[a,b]}\, \varphi_2 
&\Leftrightarrow \exists \tau \in t+[a,b]\ \text{s.t.}\ (s,\tau)\models \varphi_2\ \wedge \\
&\qquad \forall \tau' \in [t,\tau],\ (s,\tau')\models \varphi_1.
\end{aligned}
\]
}

% Additional operators are defined by:

% $\bot := \neg \top$,  
% $\varphi_1 \vee \varphi_2 := \neg (\neg \varphi_1 \wedge \neg \varphi_2)$,  
% $\mathsf{F}_{[a,b]} \varphi := \top \,\mathcal{U}_{[a,b]} \varphi$,  
% $\mathsf{G}_{[a,b]} \varphi := \neg (\mathsf{F}_{[a,b]} \neg \varphi)$.
\end{definition}
\vspace{-8 pt}
% The operators $\square$ and $\lozenge$ are called ``always'' and ``eventually'' respectively. Such operators permit to consider more tangible predicates such as “always remain within bounds” or “eventually reach a target”.

\textcolor{black}{The operators $\square$ and $\lozenge$ are called ``always'' and ``eventually,'' respectively. The operator $U_{[a,b]}$ is the (bounded) ``until'' operator. These operators allow us to express tangible predicates such as ``always remain within bounds'' or ``eventually reach a target.''}

\begin{definition}[MA-STL Formula]\label{def:mastl}
% A multi-agent Signal Temporal Logic formula $\Psi$ extends Signal Temporal Logic to multiple robots~\cite{sun2022multi}, encapsulating the desired collective behavior. Similarly, it is defined recursively:
\textcolor{black}{A multi-agent Signal Temporal Logic formula $\Psi$ lifts single-robot STL specifications to $N$ robots by composing per-robot STL formulas using team-level Boolean connectives~\cite{sun2022multi}. It is defined recursively:}
\begin{equation}
     \textcolor{black}{\Psi := \pi_i^{\varphi} \mid \neg \Psi \mid \Psi_1 \land \Psi_2 \mid \Psi_1 \lor \Psi_2,\label{eqn:phi}}
\end{equation}
where $\Psi_1, \Psi_2$ are $N$-robot STL formulas, and $\pi_i^{\varphi}$ assigns a single-robot STL specification $\varphi$ to robot $i$. Formally, the validity of an MA-STL formula with respect to trajectories $(\mathbf{s}_1, \ldots, \mathbf{s}_N)$, i.e. $\mathbf{s}_i \in S$ for $i=1,\ldots,N$,  is given by:
\textcolor{black}{
\begin{eqnarray*}
    (\mathbf{s}_1, \ldots, \mathbf{s}_N) &\models \pi_i^{\varphi} \iff (\mathbf{s}_i, 0) \models \varphi, \;  \\
    (\mathbf{s}_1, \ldots, \mathbf{s}_N) &\models\neg\Psi\Leftrightarrow\neg\big((\mathbf{s}_1, \ldots, \mathbf{s}_N)\models \Psi\big), \\
    (\mathbf{s}_1, \ldots, \mathbf{s}_N) &\models \Psi_1 \land \Psi_2 \iff (\mathbf{s}_1, \ldots, \mathbf{s}_N) \models \Psi_1 \\
    &\quad \land (\mathbf{s}_1, \ldots, \mathbf{s}_N) \models \Psi_2, \\
    (\mathbf{s}_1, \ldots, \mathbf{s}_N) &\models \Psi_1 \lor \Psi_2 \iff (\mathbf{s}_1, \ldots, \mathbf{s}_N) \models \Psi_1 \\
    &\quad \lor (\mathbf{s}_1, \ldots, \mathbf{s}_N) \models \Psi_2.
\end{eqnarray*}
}
\end{definition}

\vspace{-10 pt}

%MA-STL extends STL predicates with logical operators (and, or, not) and temporal operators (eventually, always) to define properties for the team. 

\textcolor{black}{
MA-STL assigns individual STL specifications $\pi_i^{\varphi}$ to each robot while enabling coupling through logical operators (and, or, not) and temporal operators (eventually, always).
Thus, $\Psi$ encodes both independent behaviors and collective constraints.% such as pairwise safety $\square_{[0,T]}(\|p_i(t)-p_j(t)\|_\infty > d_{\min})$.
} Each robot $ i $ has motion kinodynamic constraints:
\begin{equation}
\dot{\mathbf{x}}_i(t) = f_i(\mathbf{x}_i(t), \mathbf{u}_i(t)), \label{eq:robot_motion}
\end{equation}
where $ \mathbf{x}_i(t) $ is the state and $ \mathbf{u}_i(t) $ is the control for robot $ i $.

\begin{definition}[Kinodynamic Constraints]
Let $ \mathcal{S}_i = (f_i, X_i, U_i, \mathbf{x}_{i}^{\text{init}}) $ be a dynamical system for robot $ i $, where $ X_i \subseteq \mathbb{R}^d $ and $ U_i \subseteq \mathbb{R}^m $ are bounded state and control spaces. 
The state space $ X_i $ contains obstacles $ X_{\text{obs}} \subset X_i $, similarly the free space is defined as $ X_{i,\text{f}} = X_i \setminus X_{\text{obs}} $. 
We asume the function $ f_i: X_i \times U_i \to X_i $ to be Lipschitz, 
and $ \mathbf{x}_{i}^{\text{init}} $ is the initial state satisfying 
$ \Pi_i(\mathbf{x}_{i}^{\text{init}}) = p_i^{\text{init}}$. 
Each robot evolves according to \eqref{eq:robot_motion}, with 
$ \mathbf{x}_i(t) \in X_i $ and $ \mathbf{u}_i(t) \in U_i $. 
$ \mathbf{x}_i[\mathbf{x}_{i}^{\text{init}}, u](t) $ is the trajectory that originates at 
$ \mathbf{x}_{i}^{\text{init}}$ under control policy $ \mathbf{u}_i $. 
Let \textcolor{black}{$ V_i = U_i^{\mathbb{R}_{\geq 0}} $
}be the set of all control policies for robot $ i $.

The system $ \mathcal{S}_i $ satisfies an STL specification $ \varphi $ under a control policy $ \mathbf{u}_i \in V_i $ if the state trajectory starting at 
$ \mathbf{x}_{i}^{\text{init}} $ satisfies $ \varphi $, i.e., 
$ \mathbf{x}_i[ \mathbf{x}_{i}^{\text{init}}, u] \models \varphi $.
\end{definition}

These kinodynamic constraints ensure that planned trajectories are physically feasible, adhering to limits on velocity, acceleration, and other dynamic properties.

A finite time horizon $ T $ defines the interval $ [0,T] $ for planning and evaluation. For each robot $ i $, a positive constant $ s_i > 0 $ represents its size, modeled as an axis-aligned box centered at $ p_i(t)=\Pi_i(\mathbf{x}_i(t)) $ at time $ t $:
\begin{equation}
B_{s_i}(p_i(t)) = \{ q \in W : \|q - p_i(t)\|_\infty \leq s_i \}, \label{eq:box}
\end{equation}
where $ \| \cdot \|_\infty $ is the maximum norm. This square representation can be generalized to other shapes.

\begin{definition}[MA-STL Satisfaction with Robustness]\label{def:robust}
\textcolor{black}{Let $p_i:[0,T]\to\mathbb{R}^d$ denote the planned position trajectory of robot $i$ and $\hat{p}_i(t):=\Pi_i(\hat{\mathbf{x}}_i(t))$ its actual position trajectory induced by the executed state trajectory $\hat{\mathbf{x}}_i(t)$.}
A set of position trajectories $ (p_1(t), p_2(t), \ldots, p_N(t)) $ is 
$ \epsilon $-robust with respect to an MA-STL formula $ \Psi $ if, for any corresponding actual state trajectories 
$ \hat{\mathbf{x}}_1(t), \hat{\mathbf{x}}_2(t), \ldots, \hat{\mathbf{x}}_N(t) $ satisfying
\begin{equation}
\sup_{t \in [0, T]} \big\| \Pi_i(\hat{\mathbf{x}}_i(t)) - p_i(t) \big\| \leq \epsilon 
\quad \text{for all } i, \label{eq:tracking}
\end{equation}
and following their dynamics
\begin{equation}
\dot{\hat{\mathbf{x}}}_i(t) = f_i(\hat{\mathbf{x}}_i(t), \hat{\mathbf{u}}_i(t)) 
\quad \text{with } \hat{\mathbf{u}}_i(t) \in U_i, \label{eq:dynamic}
\end{equation}
the actual position trajectories $ (\hat{p}_1(t), \hat{p}_2(t), \ldots, \hat{p}_N(t)) \models \Phi $ and ensure no collisions, i.e., 
$ \|\hat{p}_i(t) - \hat{p}_j(t)\|_\infty \geq s_i + s_j + 2\epsilon $ for all $ i \neq j $.
\end{definition}

The parameter $ \epsilon $ ensures collision avoidance by increasing effective robot sizes and provides a robustness margin for STL satisfaction under perturbations.

\begin{problem}
Given an MA-STL motion planning problem with kinodynamic constraints 
\begin{equation*}
 % \langle W, N, T, \{p_i^{\text{init}}\}_{i=1}^N, \{s_i\}_{i=1}^N, \epsilon, \Psi, 
% \{f_i, X_i, U_i\}_{i=1}^N \rangle,   
( W, N, T, \{s_i\}_{i=1}^N, \epsilon, \Psi, \{\mathcal{S}_i\}_{i=1}^N)
% \{f_i, X_i, U_i\}_{i=1}^N
\end{equation*}
compute a set of continuous state trajectories 
$ \mathbf{x}_1(t), \mathbf{x}_2(t), \ldots, \mathbf{x}_N(t) $, 
where each $ \mathbf{x}_i(t): [0, T] \to X_i $, such that they satisfy the MA-STL  $\Psi$, respect kinodynamic constraints, ensure robustness, and guarantee collision avoidance as defined in Definitions~\ref{def:stl}–\ref{def:robust}.
\end{problem}

\section{Decoupled Robot Planning}
Here, we propose a constrained Bayesian optimization-based tree search (cBOT) algorithm for computing collision-free trajectories for individual robots. 
The cBOT planner is an extension of the BOW planner~\cite{raxit2025bow} which enjoys probabilistic completeness within a planning horizon. 
The search tree, denoted by $\mathcal{T}$, is rooted at the initial state $\mathbf{x}_\textnormal{init} \in X$ and incrementally expanded toward the goal state $\mathbf{x}_{\mathrm{goal}}$ by sequentially selecting and executing locally optimal controls. At any iteration $t$, the planner focuses on a local control neighborhood 
\begin{equation}
    V_d(\mathbf{x}_t) = \{\mathbf{u} \in U : \|\mathbf{u} - \bar{\mathbf{u}}\|_2 \leq d\},
    \label{eq:local_window}
\end{equation}
where $d > 0$ is the window radius and $\bar{\mathbf{u}}$ is the nominal control from the previous iteration. This \emph{planning window} restricts search to controls that are reachable within a short horizon, enabling both computational tractability and fine-grained maneuvering. Here, we dropped subindex $i$ from $\mathbf{x}$ for notation simplicity since this procedure applies to each robot $i$; moreover, $\mathbf{h}_t$ will denote $\mathbf{h}(t)$'s discrete version for any function $\mathbf{h}(t)$.

Within this local window, a set of $p$ candidate controls \textcolor{black}{$\{\mathbf{u}_t^{(\ell)}\}_{\ell=1}^p \subset V_d(\mathbf{x}_t)$} is sampled and evaluated using constrained Bayesian optimization (CBO). The objective function $J(\mathbf{u})$, which may encode criteria such as travel time, energy, or smoothness, is modeled as a Gaussian process (GP)
\begin{equation}
    J(\mathbf{u}) \sim \mathcal{GP}(\mu_J(\mathbf{u}), \sigma_J^2(\mathbf{u})).
    \label{eq:gp_objective}
\end{equation}
Each motion constraint $c_k(\mathbf{u}) \leq 0$, typically arising from obstacle avoidance or kinematic limits, is modeled independently as
\begin{equation}
    c_k(\mathbf{u}) \sim \mathcal{GP}(\mu_k(\mathbf{u}), \sigma_k^2(\mathbf{u})),
    \label{eq:gp_constraint}
\end{equation}
allowing the probability of constraint satisfaction to be computed as
\begin{equation}
    P_{\mathrm{feas}}(\mathbf{u}) = \prod_{k=1}^K \Phi\left(-\frac{\mu_k(\mathbf{u})}{\sigma_k(\mathbf{u})}\right),
    \label{eq:feasibility}
\end{equation}
% where $\Phi(\cdot)$ is the Gaussian cumulative distribution function.
\textcolor{black}{where $\Phi(\cdot)$ is the Gaussian cumulative distribution function and $k \in \{1,\ldots,K\}$ indexes the constraint functions $c_k$.}

The search is guided by the constrained expected improvement (CEI) acquisition function
\begin{equation}
    \mathrm{CEI}(\mathbf{u}) = \mathrm{EI}(\mathbf{u}) \cdot P_{\mathrm{feas}}(\mathbf{u}),
    \label{eq:cei}
\end{equation}
where $\mathrm{EI}(\mathbf{u})$ is the standard Bayesian optimization improvement metric over the best observed cost $J_{\mathrm{best}}$:
\begin{equation}
    \mathrm{EI}(\mathbf{u}) =
    \begin{cases}
        z(\mathbf{u}) \, \Phi(z(\mathbf{u})) + \sigma_J(\mathbf{u}) \, \phi(z(\mathbf{u})), & \sigma_J(\mathbf{u}) > 0, \\
        0, & \text{otherwise},
    \end{cases}
    \label{eq:ei}
\end{equation}
with 
\begin{equation}
    z(\mathbf{u}) = \frac{J_{\mathrm{best}} - \mu_J(\mathbf{u})}{\sigma_J(\mathbf{u})},
    \label{eq:zscore}
\end{equation}
and $\phi(\cdot)$ the Gaussian probability density function. The CEI formulation naturally balances performance improvement and safety: candidates with high $\mathrm{EI}$ but low feasibility are penalized, whereas safer but slightly less promising candidates may be favored if their overall $\mathrm{CEI}$ is higher.

Once the control
\begin{equation}
    \mathbf{u}^*_t = \underset{\mathbf{u} \in V_d(\mathbf{x}_t)}{\argmax} \;
    \mathrm{CEI}(\mathbf{u})
    \label{eq:control_selection}
\end{equation}
is selected, the state is propagated forward over a short horizon $\Delta t$ using the motion model $\dot{\mathbf{x}} = f(\mathbf{x}, \mathbf{u})$, integrated via the fourth-order Runge–Kutta method:
\begin{equation}
    \mathbf{x}_{t : t + \Delta t} = \mathrm{RK4}\left(f, \mathbf{x}_t, \mathbf{u}^*_t, \Delta t\right).
    \label{eq:rk4}
\end{equation}
The resulting short-horizon trajectory is checked for constraint satisfaction using the collision checker. If all constraints are met, the new states added to $\mathcal{T}$ as a child of $\mathbf{x}_t$ with an edge labeled by $\mathbf{u}^*_t$; otherwise, the planner selects a random valid parent from $\mathcal{T}$ and resumes from that state.
\begin{algorithm}[H]
\caption{Constrained BO-based Tree Search}
\label{alg:BOW_Planning_simple}
\begin{algorithmic}[1]
\State \textbf{Input:} Initial state $\mathbf{x}_0$, goal state $\mathbf{x}_{\mathrm{goal}}$, parameters, goal radius $r_{\mathrm{goal}}$, STL constraints $ \varphi$
\State \textbf{Output:} Motion tree, $\mathcal{T}$
\State Initialize tree with current state,  $\mathcal{T} = \{x_0\}$
\While{$\|\mathbf{x}_t - \mathbf{x}_{\mathrm{goal}}\|_\infty \leq r_{\mathrm{goal}}$}
    \State Define planning window in control space, $V_d$
	\State Initiate a dataset, $\mathcal{D} \gets \emptyset$
    \For{Sample control $\mathbf{u} \in V_d$}
        \State Evaluate cost $J(\mathbf{u})$  and constraints $c_k(\mathbf{u})$ for a short planning horizon
        \State Update dataset, $\mathcal{D} \gets D\cup\{( \mathbf{u}, J(\mathbf{u}), c_k(\mathbf{u})) \}$  
    \EndFor
    \State Train $\mathcal{GP}$ models with the dataset $\mathcal{D}$ online 
    \State Compute CEI values using equation~\eqref{eq:cei}
    \State Select control $\mathbf{u}_t^*$ with the highest CEI value using equation~\eqref{eq:control_selection}
    \State Predict states $\mathbf{x}_{t:t +\Delta t}$ by applying the same control $\mathbf{u}_t^*$ using equation~\eqref{eq:rk4}
    \If{constraints satisfy,  $\mathbf{x}_{t: t + \Delta t} \models \varphi$}
        \State Extend the motion tree $\mathcal{T}$ with  $\mathbf{x}_{t:\Delta t}$ and  $\mathbf{u}_t^*$
    \Else
        \State Update current state $\mathbf{x}_t$ by randomly sampling a valid parent state $\mathbf{x}$ from the motion tree $\mathcal{T}$
    \EndIf
\EndWhile
%\State \Return Motion tree, $\mathcal{T}$
\end{algorithmic}
\end{algorithm}

This iterative process, as summarized in Algorithm~\ref{alg:BOW_Planning_simple}, continues until the current state satisfies the goal condition $\|\mathbf{x}_t - \mathbf{x}_{\mathrm{goal}}\|_\infty \leq r_{\mathrm{goal}}$, where $r_{\mathrm{goal}}$ represents the goal tolerance radius. To guarantee collision-free trajectories, the algorithm validates each intermediate trajectory segment $\mathbf{x}_{t : t + \Delta t}$ against the set of obstacles $\mathcal{O}$ through STL specification verification, formulated as $\varphi := \bigwedge_{o \in \mathcal{O}} \square_{[0,T]}(dist(p_i(t), o) > d_\mathrm{min})$, ensuring minimum separation distances are maintained throughout the planned path. Here, $dist(p, S) = \inf_{s \in S} \|p-s\|_\infty$. Notably, this decoupled planning approach intentionally excludes inter-robot collision checking at the individual trajectory level, delegating this responsibility to the high-level STL-KCBS planner which handles multi-robot coordination and conflict resolution. 

%The synergistic combination of online Gaussian Process learning for environment adaptation, Conditional Expected Improvement-based control selection for optimal decision making, and incremental tree expansion for systematic search space exploration enables cBOT to simultaneously achieve multiple objectives: efficient exploration of the feasible solution space, refinement of local motion decisions based on learned environmental dynamics, and maintenance of safety guarantees through rigorous static obstacle avoidance.

%This iterative process, as summarized in Algorithm~\ref{alg:BOW_Planning_simple}, continues until the current state satisfies $\|\mathbf{x}_t - \mathbf{x}_{\mathrm{goal}}\|_2 \leq r_{\mathrm{goal}}$, where $r_{\mathrm{goal}}$ is the goal tolerance. To ensure that algorithm returns the sequence of states and controls defining the final collision-free trajectory, we check each intermediate trajectory $ \mathbf{x}_{t : t + \Delta t}$  with a set of $k$ static obstacles, $\mathcal{O}$, given in the form of STL specification, i.e., $\varphi:=(||x_{i}(t) - \mathcal{O}_k||_\infty > d_{min})\right)$. Note that decoupled planner intentionally avoids inter-robot collision checking which will should be dealt by the high-level STL-KCBS planner. Thus,  the combination of online GP learning, CEI-based control selection, and incremental tree expansion enables cBOT to simultaneously explore the search space, refine local motion decisions, and ensure safety guarantees with static obstacles.

\section{Signal Temporal Logic enhanced KCBS}

The Signal Temporal Logic enhanced Kinodynamic Conflict-Based Search (STL-KCBS) algorithm replaces conventional geometric intersection tests with temporal logic specifications, providing a more expressive and verifiable framework for multi-robot coordination than standard K-CBS approaches.
In terms of conflict detection, conventional K-CBS identifies conflicts through direct geometric intersection, defined as \(K = (i, j, [t_s, t_e]) \) where \( B_{s_i}(p_i(t)) \cap B_{s_j}(p_j(t)) \neq \emptyset \). In contrast, STL-KCBS employs robustness-based conflict detection using STL monitors, expressed as \( \mathcal{K} = \{(i, j, t)\in \{1,\ldots,N\}^2\times [t_s, t_e] : \mu(t) < 0\} \), where \( \mu(t) \) represents the STL robustness metric, and \( \varphi \) is the safety specification. For constraint representation, conventional K-CBS generates spatial-temporal constraints such as \( C_i = \{ (\mathbf{x}, t)\in X_i\times [t_s, t_e]  : \mathbf{x} \in B_{s_i}(p_i(t)) \} \). However, STL-KCBS expresses constraints as temporal logic formulas that can encode complex behavioral requirements beyond simple collision avoidance.

The STL-KCBS algorithm enhances the conventional K-CBS framework by integrating STL monitors into the conflict detection mechanism, offering several theoretical and practical advantages for multi-robot motion planning. At its core, the algorithm initializes an STL safety monitor \( \varphi \) for each robot pair \( (i, j) \) where $i\neq j$. This monitor encodes safety specifications that efficiently enforce collision avoidance, including spatio-temporal safety constraints such as \( \varphi = \square_{[0,T]} (\|p_{i}(t) - p_{j}(t)\|_\infty > d_{\textnormal{min}}) \). 

% , velocity-dependent safety margins like \( \varphi_{dynamic} = \square_{[0,T]} (||p_{p_1} - p_{p_2}|| > d_{min} + \alpha \cdot ||v_{rel}||) \), and temporal coordination requirements such as \( \varphi_{coord} = \square_{[0,T]} (\text{at\_intersection}(p_1) \rightarrow \neg \Diamond_{[0,\tau]} \text{at\_intersection}(p_2)) \).

%Robustness-based conflict detection in STL-KCBS utilizes the STL robustness metric \( \mu(t) \), which provides a quantitative measure of specification satisfaction: greater than zero if \( \varphi \) is satisfied with margin, equal to zero if marginally satisfied, and less than zero if violated. This continuous metric enables gradient-based optimization for trajectory refinement, early conflict prediction through robustness degradation monitoring, and quantitative safety margins for robust planning.
\textcolor{black}{
Robustness-based conflict detection in STL-KCBS utilizes the STL robustness metric $\mu(t)$, which generalizes geometric intersection tests by evaluating distance-based predicates over time. 
The metric is positive if $\varphi$ is satisfied with margin, zero if marginal, and negative if violated, enabling early conflict prediction and quantitative safety margins for robust planning.}
 At each timestep \( t \), the algorithm constructs a comprehensive state vector 
\( \Bar{\mathbf{x}} = [t, {p_i}(t), {p_j}(t)]^T \) to evaluate the STL specification. 
% \( \mu(t) = [p_{p_1}(t), v_{p_1}(t), a_{p_1}(t), p_{p_2}(t), v_{p_2}(t), a_{p_2}(t), \text{env}(t)]^T \), incorporating positions, velocities, accelerations, and environmental context to support rich temporal specifications.

\begin{algorithm}[htb]
\caption{Kinodynamic Conflict-Based Search with STL Monitor} \label{alg:stl_kcbs}
\begin{algorithmic}[1]

\Procedure{STLConflictSearch}{$plan$}
    \State $T \gets \max(\text{trajectory lengths in plan})$
    \State $\text{conflicts} \gets \emptyset$
    \For{each robot pair $(i, j)$}
       \State Initialize STL safety monitor $\mu$ for $p_i$ and $p_j$
        % \Comment{Populate STL monitor with robot trajectories}
        \For{$t = 0$ to $T - 1$}
            \State $\Bar{\mathbf{x}} \gets \text{create state vector at timestep } t$
            \State $\mu.\text{add\_sample}(\Bar{\mathbf{x}})$
        \EndFor

        \For{$t = 0$ to $T - 1$}  
            \If{$\mu(t) < 0$}
                \State $\text{conflicts} \gets \text{conflicts} \cup \{(i, j, t)\}$ \\
                \Comment{Add conflict between $i$ and $j$ at timestep $t$}
            \EndIf
        \EndFor
    \EndFor    
    \State \Return $\text{conflicts}$
\EndProcedure
\end{algorithmic}
\end{algorithm}

Algorithm~\ref{alg:stl_kcbs} shows the workflow of STL-KCBS conflict detection which begins with an initialization phase, where the maximum trajectory length \( T \) across all robots is determined, and STL monitors are initialized for each robot pair with appropriate safety specifications. In the monitor population step, for each timestep \( t \in [0, T-1] \), the state vector \( \Bar{\mathbf{x}} \) is constructed from a pairwise robot trajectories, and samples are added to the STL monitor via \( \mu.\text{add\_sample}(\Bar{\mathbf{x}}) \). During robustness evaluation, for each timestep \( t \), the robustness \( \mu(t) \) is computed; if \( \mu(t) < 0 \), a conflict \( (i, j, t) \) is recorded. Finally, conflict resolution involves generating STL-constrained trajectory refinements and propagating temporal constraints through the search tree.

Implementation considerations for STL-KCBS include hierarchical decomposition with careful monitor design, where the choice of STL specifications impacts performance; longer horizons increase expressiveness but raise computational costs, atomic predicates must balance granularity with evaluation efficiency, and robustness computation can leverage efficient algorithms. Integration with low-level planning requires the planner to be STL-aware, minimizing the cost-to-go function, i.e., 
% $\min_{\mathbf{u}(t)} \| f(\mathbf{x}(t), \mathbf{u}(t))  - \mathbf{x}_{\mathrm{goal}}\|_2$ subject to robust satisfaction, \( \mu(t) \geq 0 \).
$\min_{u_i\in U_i} \| \mathbf{x}_i^\mathrm{init} + \int_{0}^{T} f_i(\mathbf{x}_i(t), \mathbf{u}_i(t)) dt  - \mathbf{x}_i^{\mathrm{goal}}\|_2$ subject to robust satisfaction, \( \mu(t) \geq 0 \). 
\section{Experiments and Results}

% All experiments were conducted on a desktop computer running Ubuntu22.04 LTS, equipped with an Intel(R) Core(TM) i7-10700 CPU @ 2.90GHz, and 32,GB of RAM. Indoor robot localization was provided by a VICON motion capture system operating at 120 Hz where the outdoor localization was provided by standard GPS sensors. The decoupled planner was configured with a maximum velocity of 1.0\,m/s, a minimum velocity of 0.0\,m/s, a maximum acceleration of 0.5\,m/s\textsuperscript{2}, a maximum yaw rate of 0.6981\,rad/s, and a maximum yaw acceleration of 2.0472\,rad/s\textsuperscript{2}. The prediction horizon ranging from 3.0\,s to 50.0\,s where $dt$ was set to 0.1. For KCBS, the merge bound parameter was selected within the range of 25 to 60. For trajectory optimization within the decoupled planner, we employed a Squared Exponential ARD kernel (Radial Basis Function) to efficiently learn local cost map.
All experiments were performed on a desktop computer running Ubuntu 22.04 LTS with an Intel® Core™ i7-10700 CPU operating at 2.90 GHz and 32 GB of RAM. For robot localization, we utilized a VICON motion capture system running at 120 Hz for indoor environments and standard GPS sensors for outdoor scenarios. The decoupled planner operated under the following kinodynamic constraints: maximum velocity of 1.0 \si{\meter\per\second}, minimum velocity of 0.0 \si{\meter\per\second}, maximum acceleration of 0.5 \si{\meter\per\second^2}, maximum yaw rate of 0.6981 \si{\radian\per\second}, and maximum yaw acceleration of 2.0472 \si{\radian\per\second^2}. We configured the prediction horizon between 3.0 \si{\second} and 50.0 \si{\second} with a time discretization step of 0.1 \si{\second}. The KCBS algorithm employed merge bound parameters ranging from 25 to 60. For GP kernel within the decoupled planner, we utilized a Squared Exponential ARD kernel (Radial Basis Function) to effectively learn the local cost map.

%\subsection{STL Specification and Hierarchical Decomposition}
%Here, we combine the decoupled planner cBOT and the high-level planner STL-KCBS. We refer this planner as STLcBOT.   
%The STLcBOT accepts safety constraints formulated as Signal Temporal Logic (STL) specifications for both inter-robot and obstacle avoidance. The safety constraints are expressed as: 
%\begin{equation}
%\begin{split}
%\varphi_{safety} =& \square_{[0,T]} \left((||x_{i}(t) - x_{j}(t)||_\infty > d_{min}) \right. \\
%&\left. \wedge (||x_{i}(t) - \mathcal{O}_k||_\infty > d_{min})\right),
%\end{split}
%\end{equation}
%ensuring minimum separation distances between robots and obstacles throughout the mission duration, while reachability constraints are formulated as $\varphi_{reachability} = \diamond_{[0,T]} (||x_{i}(t) - x_{g_i}|| < r_{goal})$, guaranteeing that each robot reaches its designated goal region. The complete mission specification $\varphi$ combines these constraints through logical conjunction to ensure both safety and task completion such that $\varphi := \varphi_{safety} \wedge \varphi_{reachability}$. To efficiently synthesize multirobot trajectories under $\varphi$ constraints, we performed hierarchical decomposition where the cBOT planner is responsible to satisfy reachability and static obstacle avoidance objectives, on the other hand, the STL-KCBS is responsible for inter-robot collision avoidance objective through pair-wise conflict-based search technique.  

\subsection{STL Specification and Hierarchical Decomposition}
STLcBOT combines the decoupled cBOT planner with the high-level STL-KCBS planner to handle multi-robot coordination under STL constraints. Safety constraints ensure minimum separation distances throughout the mission:
\begin{equation*}
\begin{split}
\varphi_{\mathrm{safety}} &=\bigwedge_{i \neq j }  \Big(\square_{[0,T]}\big(||p_{i}(t) - p_{j}(t)||_\infty > d_\mathrm{min}\big)\Big) \\
&\wedge \bigwedge_{o \in \mathcal{O}} \big( dist(p_i(t), o) > d_\mathrm{min}\big),
\end{split}
\end{equation*}
while reachability constraints guarantee goal achievement: $\varphi_\mathrm{reachability} = \lozenge_{[0,T]} (\| p_{i}(t) - p_i^{\mathrm{goal}}\|_\infty < r_\mathrm{goal})$. The complete mission specification combines both constraints: $\varphi := \varphi_\mathrm{safety} \wedge \varphi_\mathrm{reachability}$. The hierarchical decomposition assigns cBOT to handle reachability and static obstacle avoidance, while STL-KCBS manages inter-robot collision avoidance through pairwise conflict resolution.

\begin{figure}[t]
    \centering
    \begin{subfigure}[b]{0.23\textwidth}
        \centering
        \includegraphics[width=\textwidth]{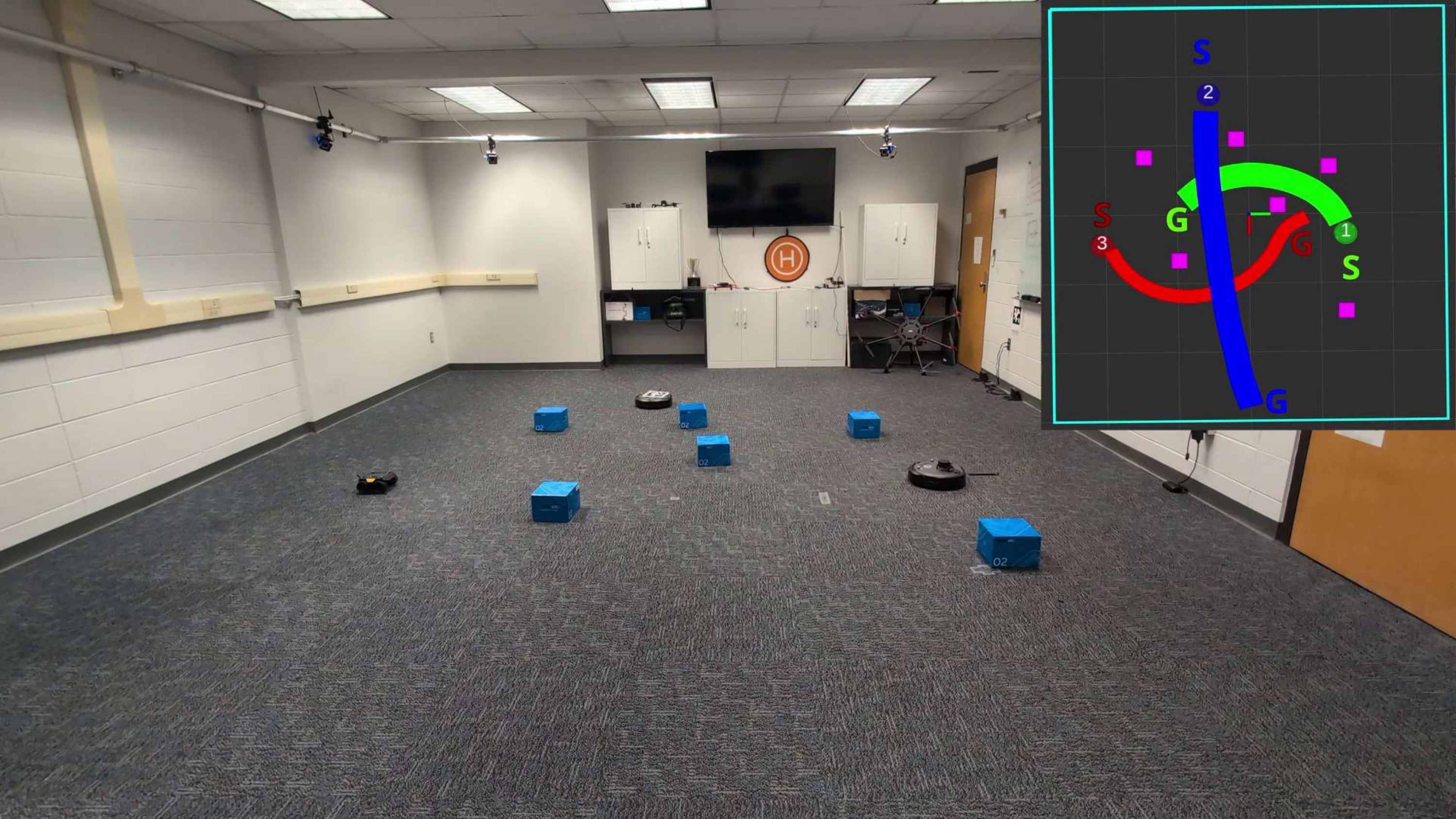}
        \caption{}
        \label{subfig:gv_e_1}
    \end{subfigure}
    \begin{subfigure}[b]{0.23\textwidth}
        \centering
        \includegraphics[width=\textwidth]{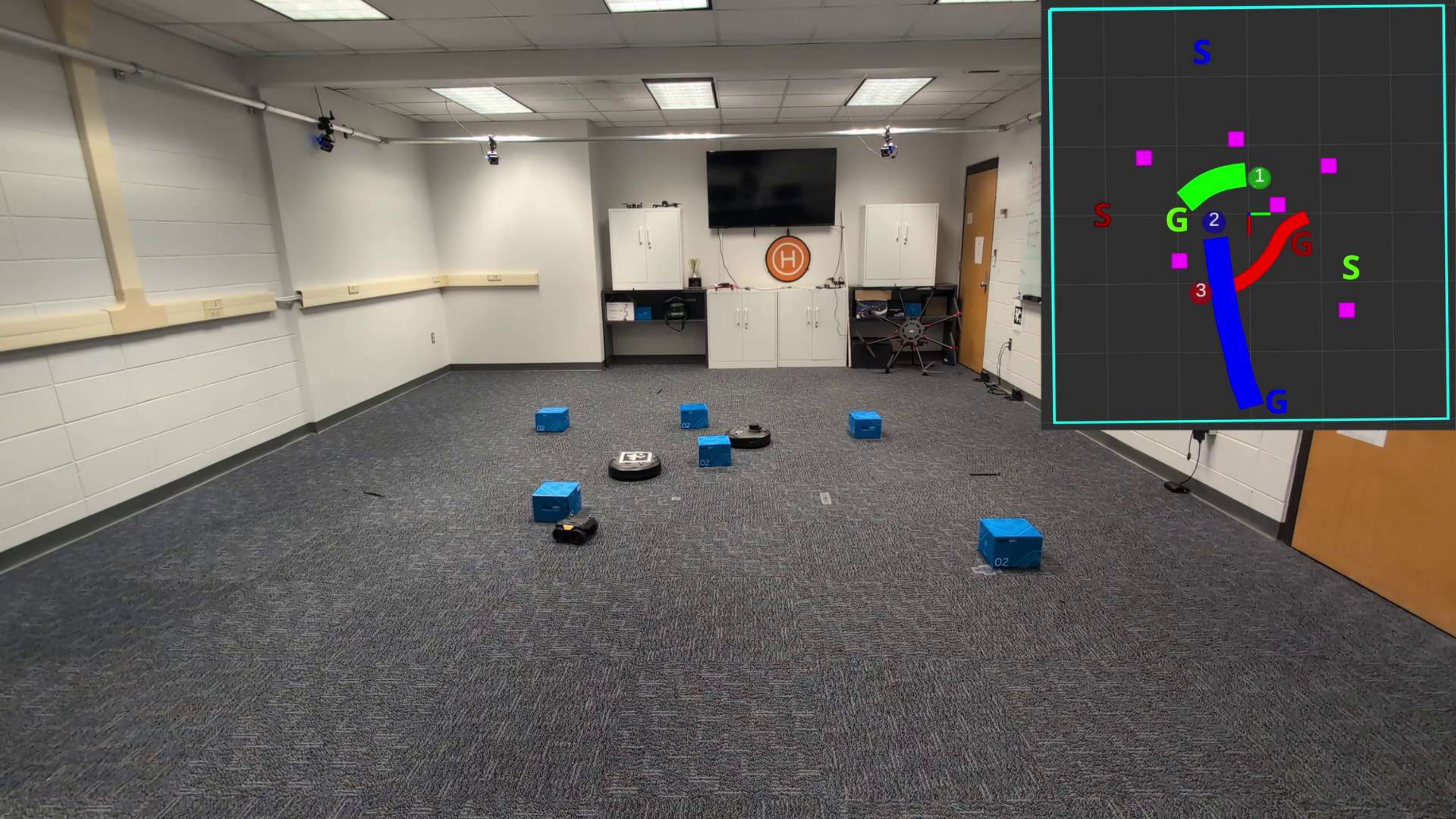}
        \caption{}
        \label{subfig:gv_e_2}
    \end{subfigure}
    
    \caption{STLcBOT planner applied to three UGVs navigating a cluttered indoor environment. Subfigures~\ref{subfig:gv_e_1} and~\ref{subfig:gv_e_2} demonstrate kinodynamically-constrained trajectories that enable accurate trajectory following and collision avoidance.}
    \label{fig:gv_exp}
\end{figure}

\begin{figure}[t]
    \centering
    \begin{subfigure}[b]{0.22\textwidth}
        \centering
        \includegraphics[width=\textwidth]{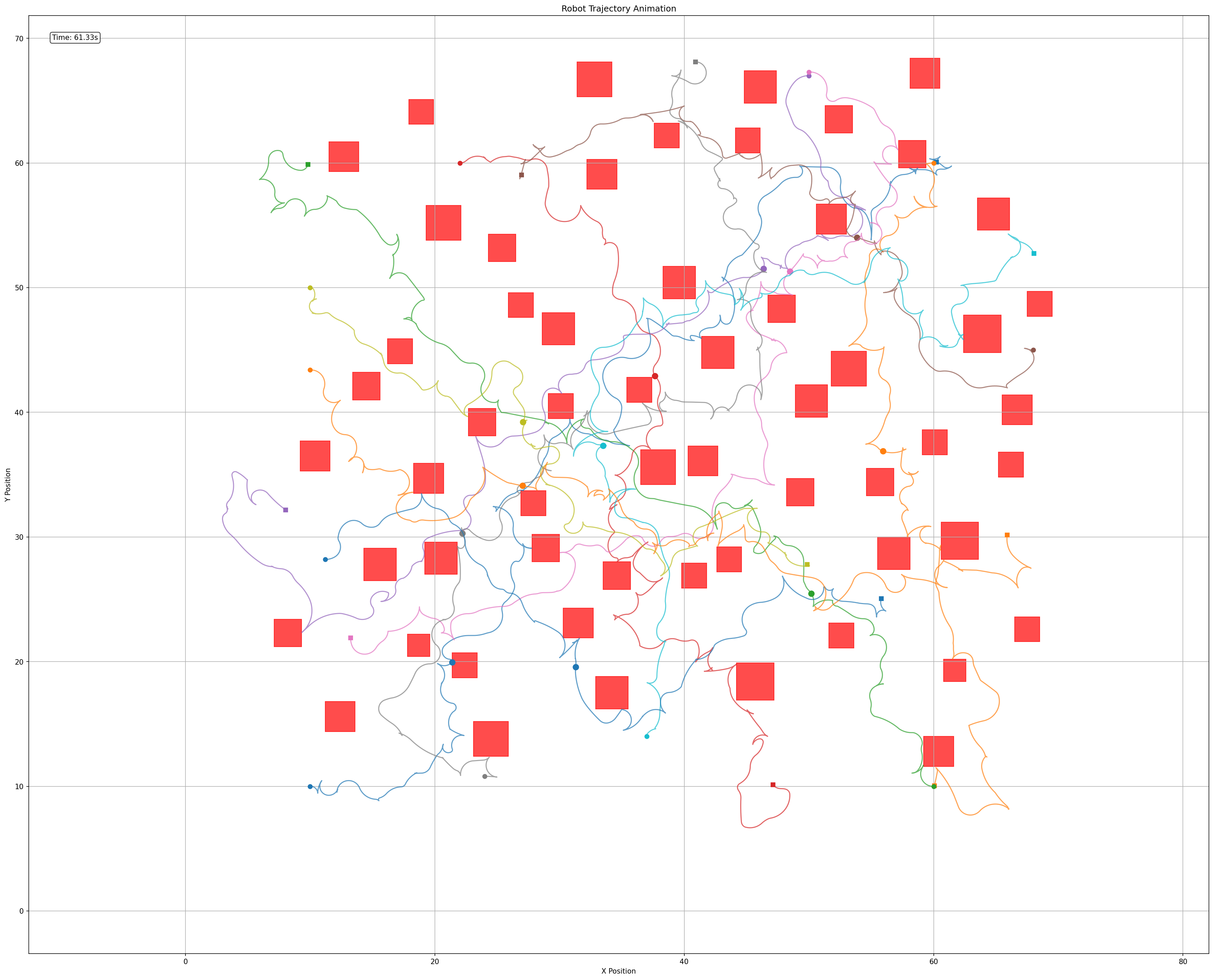}
        \caption{}
        \label{subfig:sub1}
    \end{subfigure}
    \hfill
    \begin{subfigure}[b]{0.22\textwidth}
        \centering
        \includegraphics[width=\textwidth]{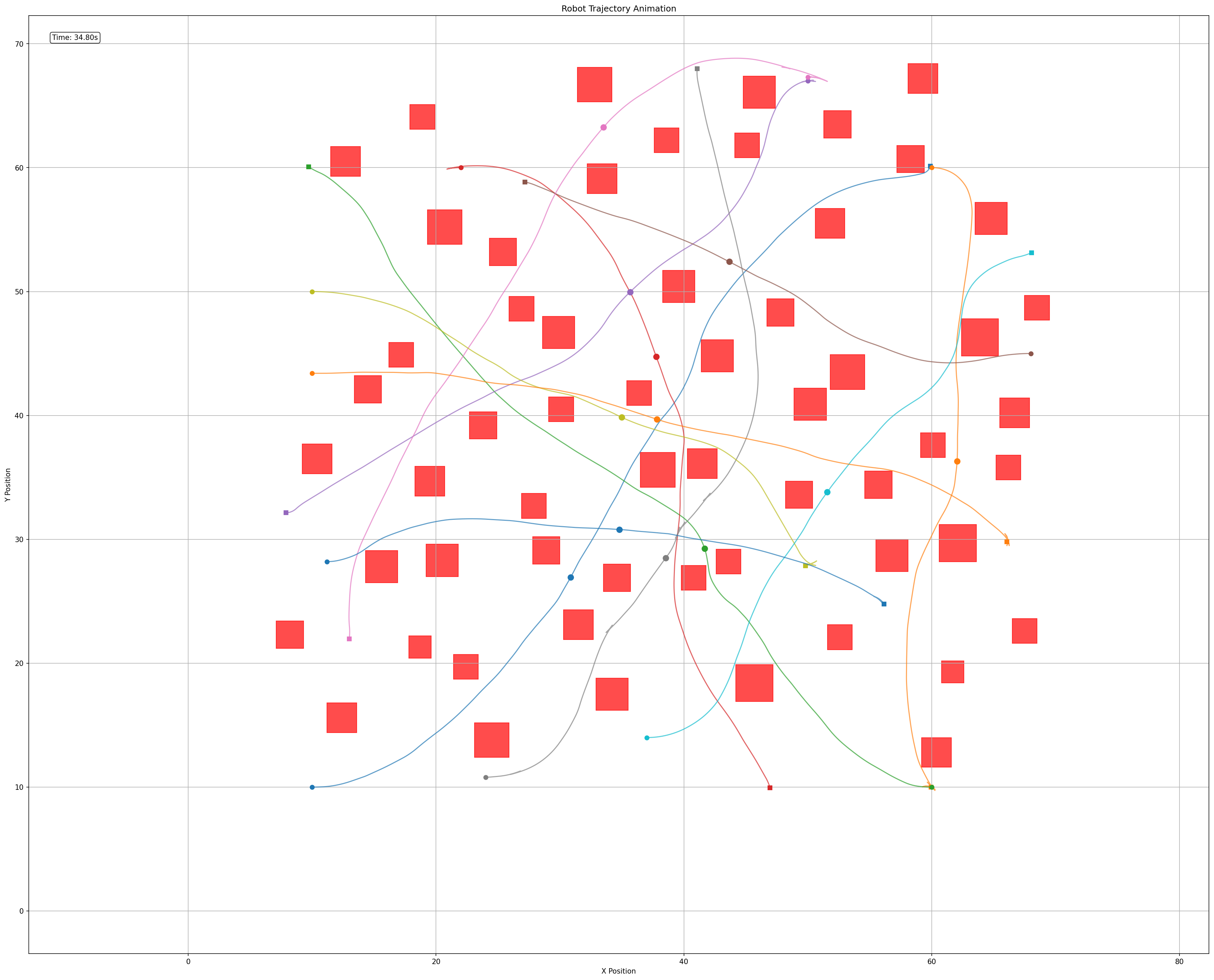}
        \caption{}
        \label{subfig:sub2}
    \end{subfigure}
    
    \vspace{0.5em} % space between rows
    
    \begin{subfigure}[b]{0.22\textwidth}
        \centering
        \includegraphics[width=\textwidth]{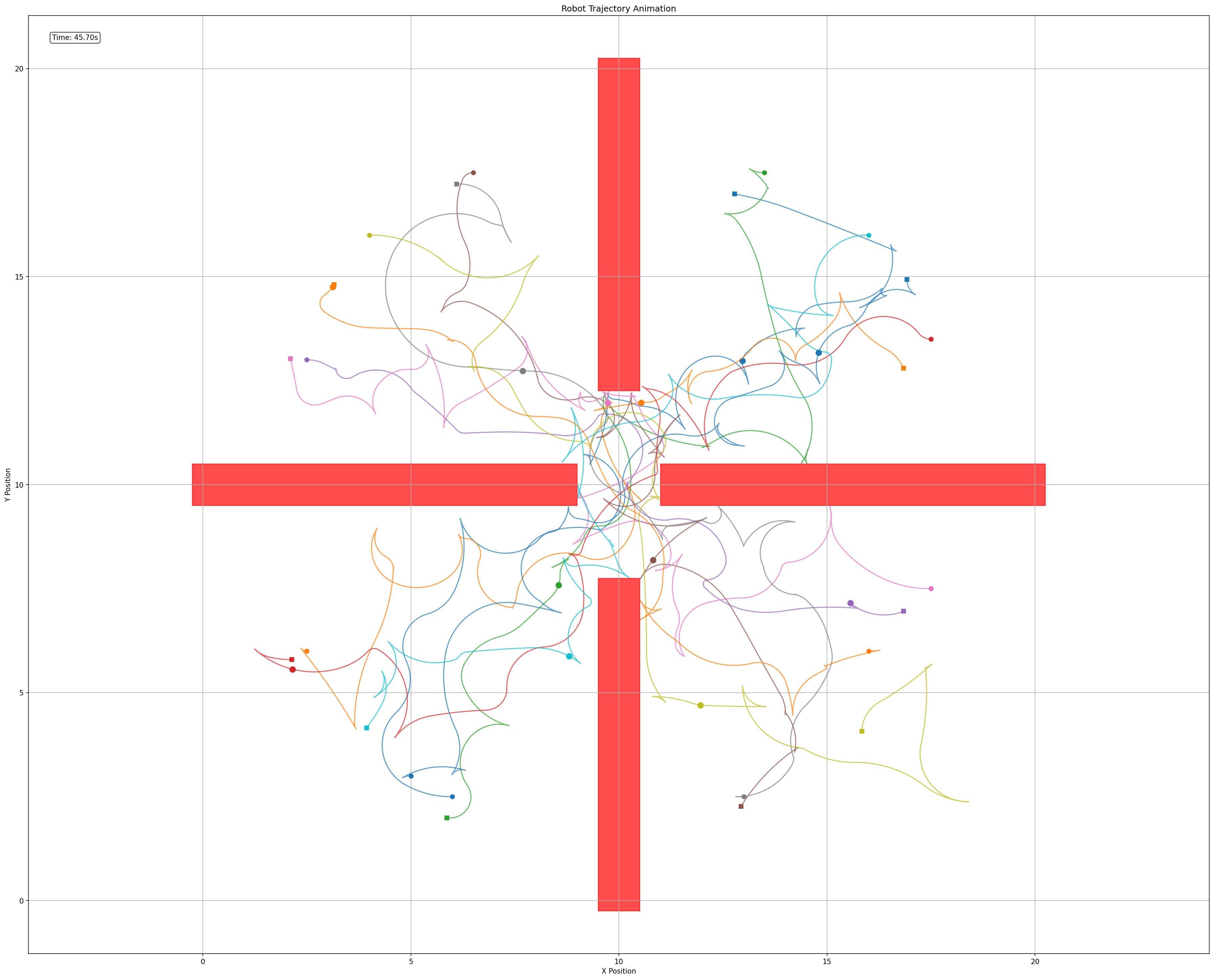}
        \caption{}
        \label{subfig:sub3}
    \end{subfigure}
    \hfill
    \begin{subfigure}[b]{0.22\textwidth}
        \centering
        \includegraphics[width=\textwidth]{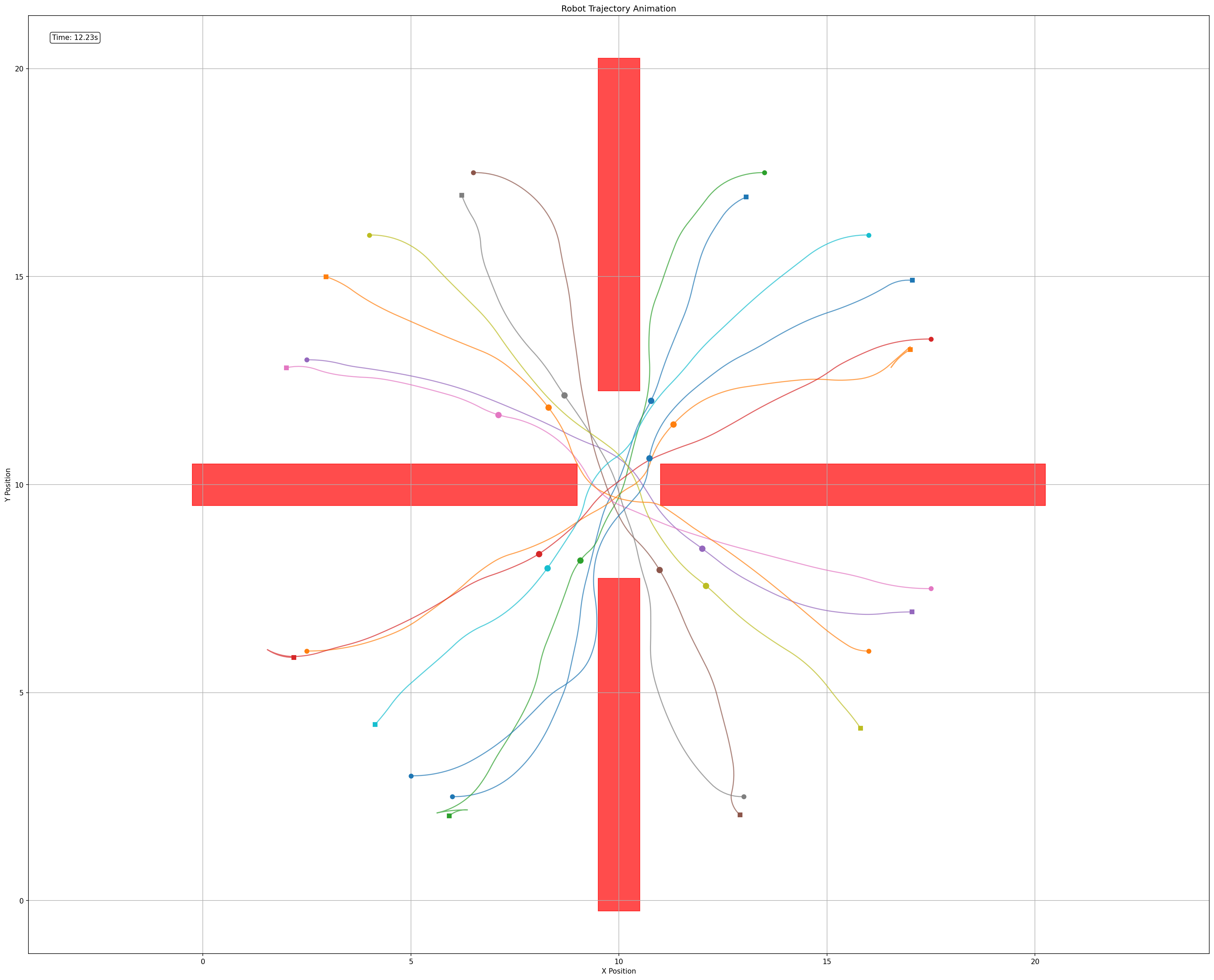}
        \caption{}
        \label{subfig:sub4}
    \end{subfigure}
    
    \caption{Comparison of RRT and cBOT planners for twelve robots under STL constraints in forest and cross-hall environments. Figures~\ref{subfig:sub1} and \ref{subfig:sub3} show RRT-generated trajectories, while Figures~\ref{subfig:sub2} and \ref{subfig:sub4} display cBOT-generated paths. cBOT produces notably shorter and smoother trajectories compared to the longer, less structured paths from RRT-based planning.}
    \label{fig:rrt_bow}
\end{figure}

\subsection{Indoor Experiments and Qualitative analysis}

We conducted controlled indoor experiments using a heterogeneous multi-robot system comprising two iRobot Create~3 educational robots (radius $0.17$ \si{\meter} each) and one Waveshare Rover robot ($0.19$ \si{\meter} width $\times$ $0.17$ \si{\meter} width) operating within a $6.5$ \si{\meter} $\times$ $5.5$ \si{\meter} bounded workspace containing six static obstacles ($0.2$ \si{\meter} $\times$ $0.2$ \si{\meter} each) at known positions, denoted by $\mathcal{O}=\{o_1,\ldots,o_6\}$.  As illustrated in Fig.~\ref{fig:gv_exp}, three color-coded robots were tasked with navigating from initial positions $(0.50, 1.35)$ \si{\meter}, $(-1.87, -0.57)$ \si{\meter}, and $(0.16, -2.00)$ \si{\meter} to goal locations $(-0.18, -0.9)$ \si{\meter}, $(2.8, 0.1)$ \si{\meter}, and $(0.05, 0.82)$ \si{\meter} respectively, while avoiding the magenta-colored static obstacles. The STLcBOT solver successfully computed collision-free trajectories within \textcolor{black}{$42$~\si{\milli\second}}, generating paths of lengths $2.933$ \si{\meter}, $4.73$ \si{\meter}, and $3.84$ \si{\meter} for the three robots with corresponding average velocities of $0.16$ \si{\meter\per\second}, $0.18$ \si{\meter\per\second}, and $0.16$ \si{\meter\per\second}, totaling $11.503$ \si{\meter} in combined path length. Despite spatial intersections in the planned paths, temporal coordination through differentiated control inputs ensured collision-free execution throughout the mission. The approach leverages the cBOT planner for individual trajectory generation while the STL-enhanced K-CBS planner employs merge and restart heuristics for efficient conflict validation, achieving high success rates in cluttered environments as demonstrated in the accompanying project website video.

To further evaluate the effectiveness of our approach, Fig.~\ref{fig:rrt_bow} presents a qualitative comparison between STLcBOT and RRT-based planners in two challenging environments: a Poisson forest and a cross-hall configuration. The comparison reveals two key qualitative advantages of our method. First, trajectory smoothness is significantly improved, as cBOT generates visually smoother paths compared to the characteristically jagged trajectories produced by RRT algorithms. Second, trajectory efficiency is enhanced through shorter path lengths, demonstrating superior navigation performance. The contrast between Figs~\ref{fig:rrt_bow}a and \ref{fig:rrt_bow}c versus Figs~\ref{fig:rrt_bow}b and \ref{fig:rrt_bow}d clearly illustrates that RRT produces longer and less structured paths, while cBOT yields more organized and coordinated motion patterns. This visual difference is particularly pronounced in the cross-hall environment, where the structured nature of cBOT trajectories becomes evident. 

\subsection{Outdoor Experiments in a Lake environment}

We conducted field experiments in a lake environment using fleets of two and three Autonomous Surface Vehicles (ASVs), each equipped with YSI EXO2 sonde sensors for measuring water quality parameters. The experimental workspace encompassed a $25$\,m $\times$ $35$\,m lake area containing two circular fountain obstacles with $2$\,m radii. Localization was achieved through sensor fusion of GPS and IMU data, providing accurate position estimates for trajectory tracking. For the two-ASV configuration, we designed three challenging scenarios that tested the planner's capability to handle complex spatial coordination: X-pattern navigation where both vehicles traverse intersecting paths with high point density in the central lake region, and position-swapping maneuvers where ASV 1's initial position served as ASV 2's goal location and vice versa. The three-ASV experiment introduced additional complexity by requiring ASV 3 to navigate to a goal location positioned at the intersection of the planned trajectories for ASV 1 and ASV 2, creating a multi-vehicle coordination challenge that demands precise temporal and spatial planning to avoid collisions while maintaining mission objectives. 
\begin{figure}[t]
    \centering
    \begin{subfigure}[b]{0.15\textwidth}
        \centering
        \includegraphics[width=\textwidth,height=0.95\textwidth]{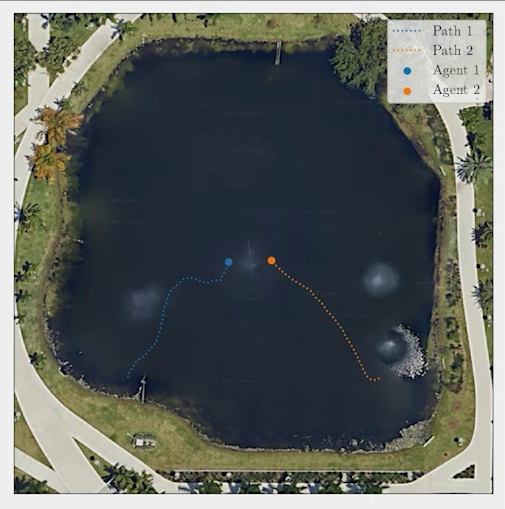}
        \caption{}
        \label{subfig:asv_1}
    \end{subfigure}
    \begin{subfigure}[b]{0.15\textwidth}
        \centering
        \includegraphics[width=\textwidth,height=0.95\textwidth]{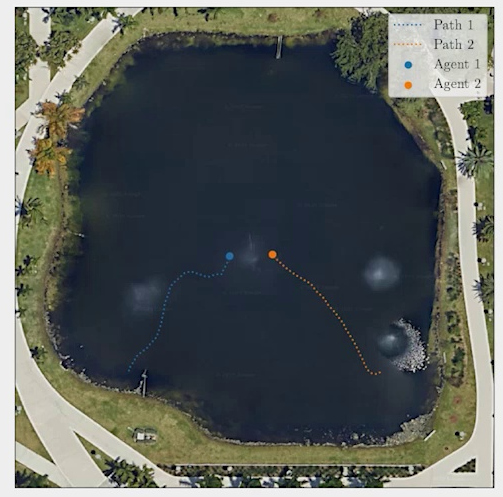}
        \caption{}
        \label{subfig:asv_2}
    \end{subfigure}
        \begin{subfigure}[b]{0.15\textwidth}
        \centering
        \includegraphics[width=\textwidth,height=0.95\textwidth]{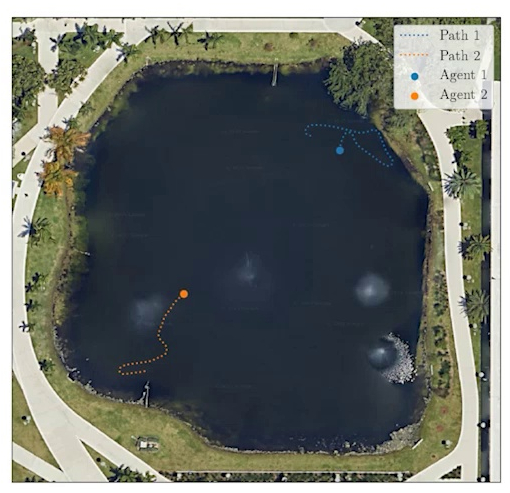}
        \caption{}
        \label{subfig:asv_3}
    \end{subfigure}
    
    \vspace{0.5em} % space between rows
    
    \begin{subfigure}[b]{0.15\textwidth}
        \centering
        \includegraphics[width=\textwidth]{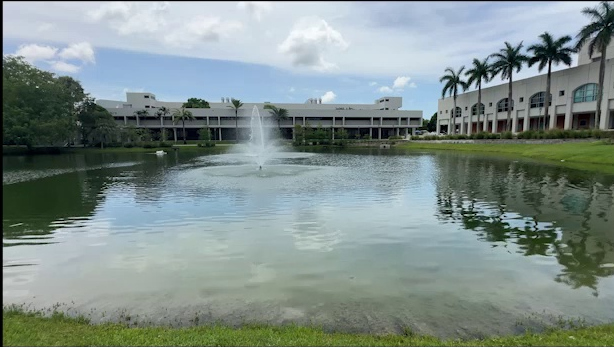}
        \caption{}
        \label{subfig:asv_4}
    \end{subfigure}
    \begin{subfigure}[b]{0.15\textwidth}
        \centering
        \includegraphics[width=\textwidth]{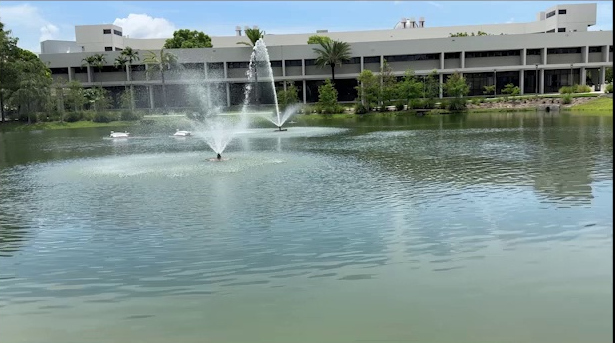}
        \caption{}
        \label{subfig:asv_5}
    \end{subfigure}
        \begin{subfigure}[b]{0.15\textwidth}
        \centering
        \includegraphics[width=\textwidth]{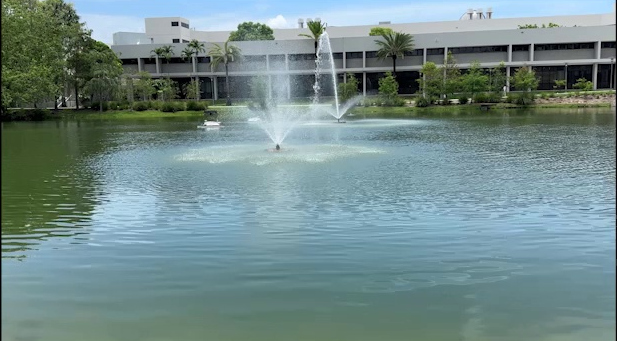}
        \caption{}
        \label{subfig:asv_6}
    \end{subfigure}
    
    \caption{Field validation of STLcBOT using two ASVs in a lake environment with complex navigation scenarios. Figs.\subref{subfig:asv_1}--\subref{subfig:asv_3} show planned trajectories, while Figs.\subref{subfig:asv_4}--\subref{subfig:asv_6} display ASVs executing missions in the operational area.}
    \label{fig:field_asv}
\end{figure}

\begin{figure}[t]
    \centering
    \begin{subfigure}[b]{0.23\textwidth}
        \centering
        \includegraphics[width=\textwidth]{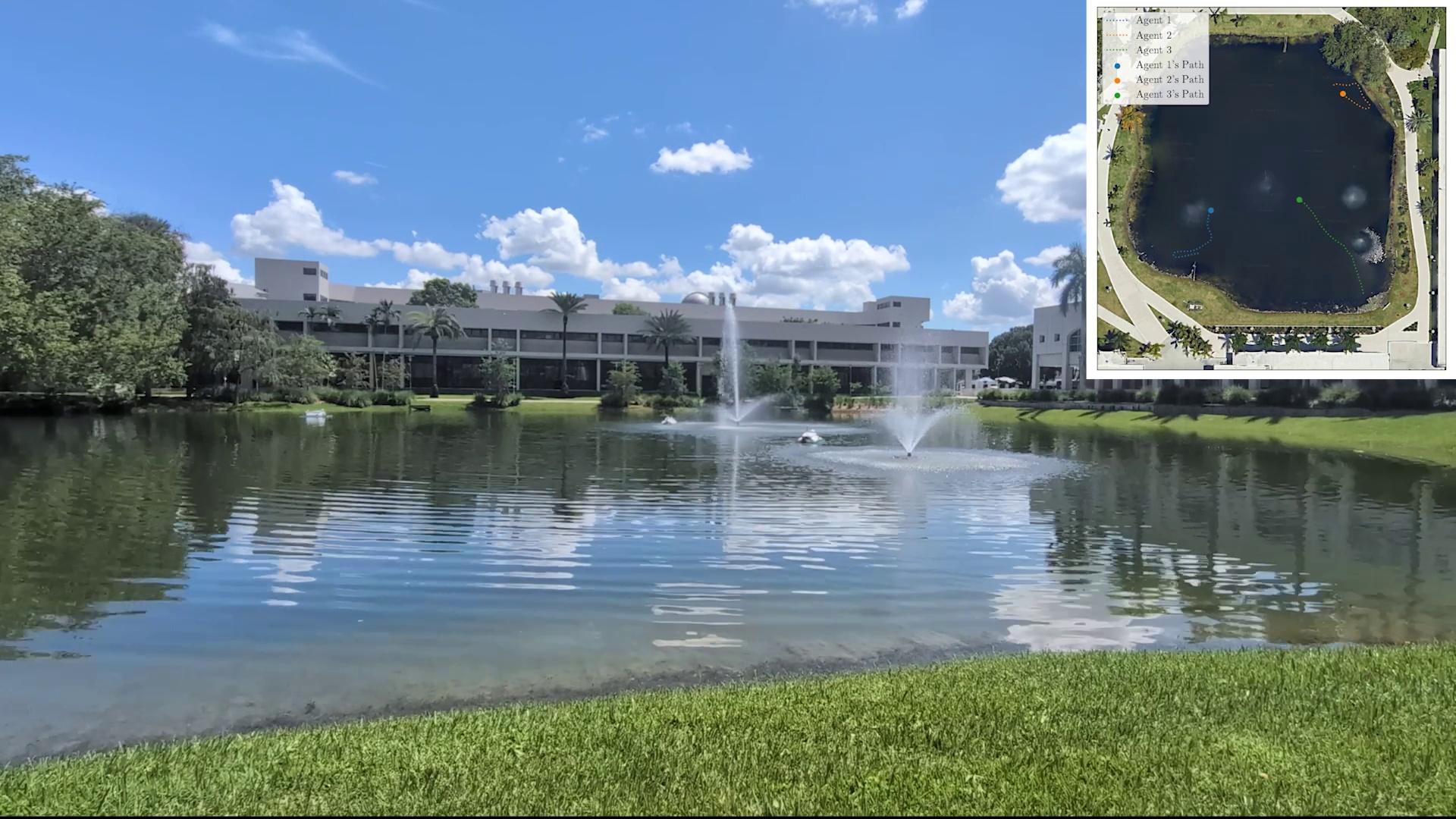}
        \caption{}
        \label{subfig:asv_3_1}
    \end{subfigure}
    \begin{subfigure}[b]{0.23\textwidth}
        \centering
        \includegraphics[width=\textwidth]{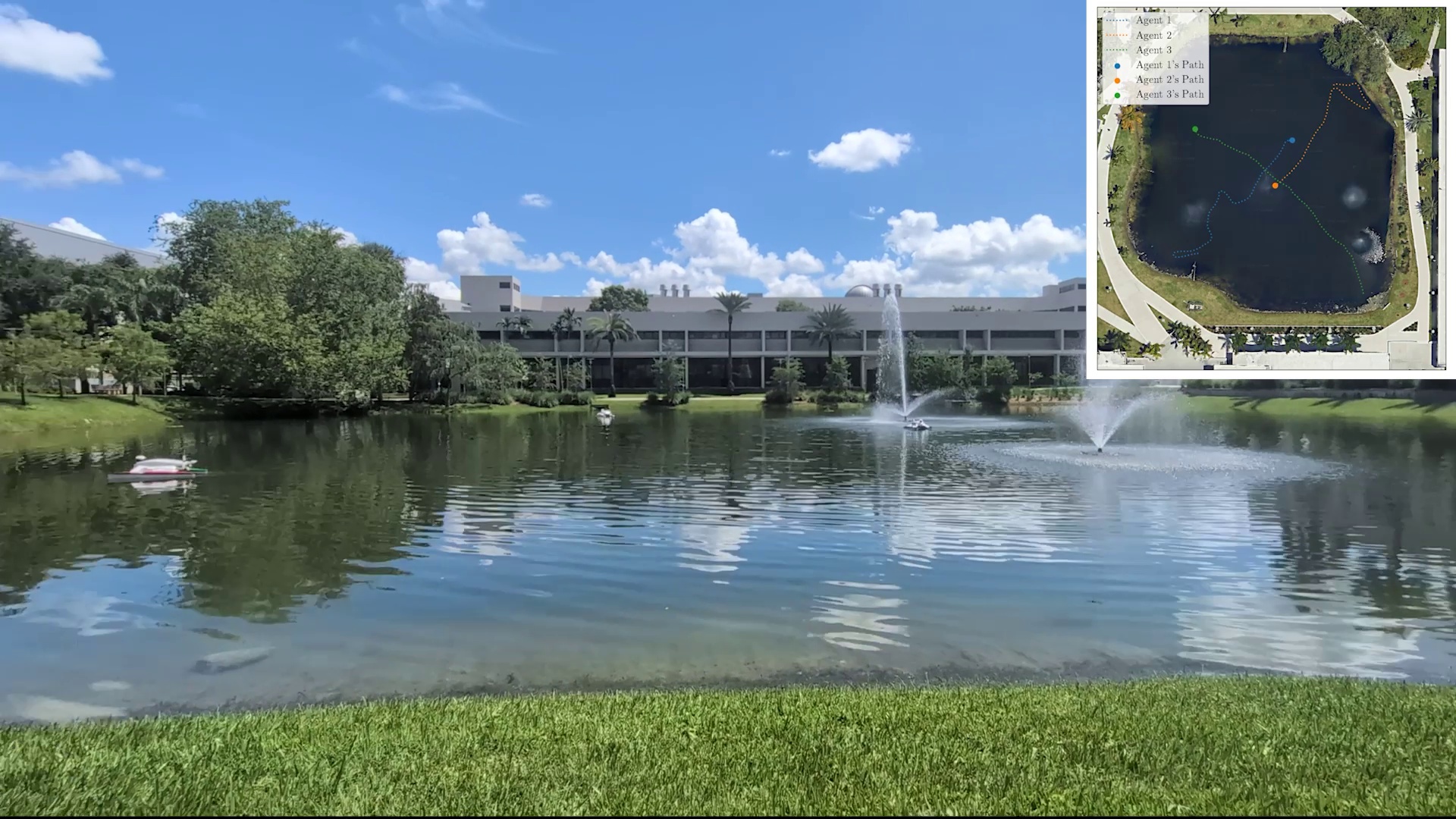}
        \caption{}
        \label{subfig:asv_3_2}
    \end{subfigure}
    
    \caption{STLcBOT planner applied to three ASVs navigating in a lake environment with challenging spatial configurations. Subfigures (\ref{subfig:asv_3_1}) and (\ref{subfig:asv_3_2}) show trajectory execution where ASVs reach their respective goals while avoiding fountain obstacles.}
    \label{fig:asv_3_exp}
\end{figure}

In all cases, STLcBOT computed collision-free trajectories in under one second. The three robots achieved average path lengths of $80.10$ \si{\meter}, $73.82$ \si{\meter}, and $72.62$ \si{\meter}, with corresponding average velocities of $0.1676$ \si{\meter\per\second}, $0.1651$ \si{\meter\per\second}, and $0.1669$ \si{\meter\per\second}, totaling $226.54$ \si{\meter}. Despite lower localization accuracy compared to indoor tests and intersecting trajectories, differentiated control inputs enabled collision-free execution. These results demonstrate STLcBOT’s ability to handle heterogeneous vehicles and their kinodynamic constraints in real-world aquatic environments.
%
% to follow the trajectories generated by STBOW within a $25 \times 35$ \si{\meter}eter workspace at our campus lake. During each run, measurements were recorded at 1 Hz, producing a dense spatiotemporal dataset for evaluating mapping accuracy and the consistency of multi-robot sampling in a confined aquatic environment. The ASVs were localized by combining GPS and IMU data, while DMRIG coordinated their motion to enable efficient sampling, maintain coverage, and actively avoid inter-vehicle collisions. The algorithm’s performance was validated by the successful generation of an online temperature map, as shown in Figure \ref{fig:field_experiment}.
%Each mission lasted approximately 6–7 minutes (390–400 s), during which the robots executed trajectories designed to maximize coverage. The average path length was $\approx$ 155 m. In particular, ASV1 covered a path length of about 167 m, while ASV2 and ASV3 followed slightly shorter survey trajectories of 141m and 156m, respectively. The distribution of path lengths was balanced, with $\approx 8\%$ variation across vehicles due to differences in sampling roles assigned by DMRIG. This illustrates the flexibility of DMRIG to dynamically balance local exploration with global coordination. Temperature measurements were consistent across robots, with mean value of $31.4^\circ$C across the workspace.

\begin{figure*}[h]
    \centering
    \includegraphics[width=0.75\textwidth]{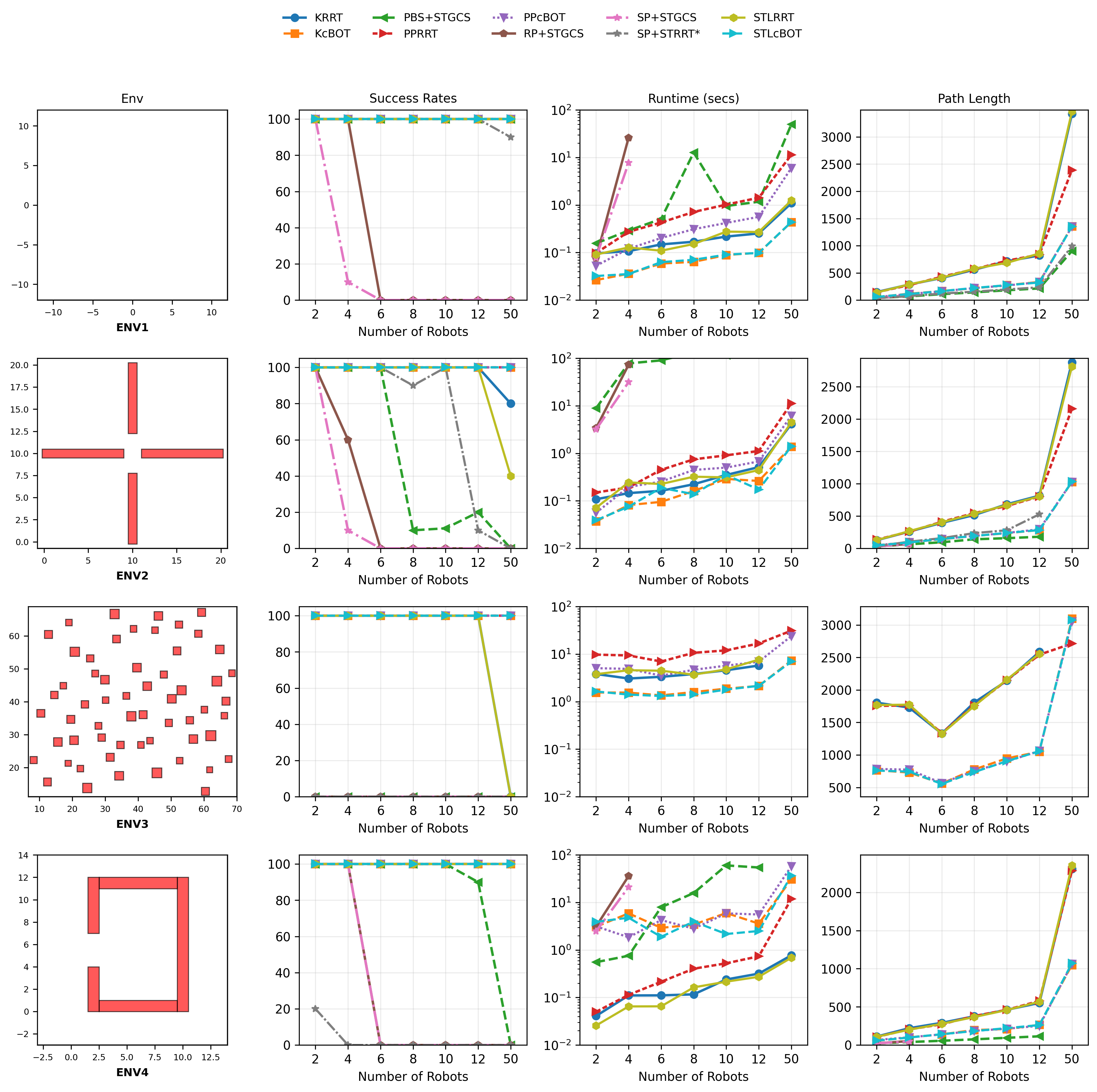}
    \caption{Benchmark comparison of multi-robot motion planning algorithms across four representative environments (Env.~1: empty, Env.~2: cross-hall, Env.~3: forest, Env.~4: bugtrap). 
    The columns report success rates, runtimes, and path lengths as the number of robots increases. 
    The proposed STLcBOT, together with other cBOT-based methods (KcBOT, PPcBOT)~\cite{kottinger2022conflict,ma2019searching}, consistently achieves near-perfect success and efficient paths with tractable runtimes. 
    RRT-based methods (PPRRT, STLRRT, KRRT)~\cite{lavalle2001randomized,kottinger2022conflict,ma2019searching,maler2004monitoring} degrade under clutter and density, while convex optimization–based approaches (PBS+STGCS, SP+STGCS, RP+STGCS)~\cite{tang2025space} and temporal sampling extension (SP+STRRT*)~\cite{tang2025space} generate competitive paths when successful but fail to scale, particularly in Env.~3.}
    \label{fig:bench}
    \vspace{-17pt}
\end{figure*}

\subsection{Benchmark Results}

We evaluate the proposed STLcBOT framework against nine baseline planners. All methods were executed for 10 independent trials under identical conditions. The baselines span a diverse set of planning paradigms and implementations. STLcBOT (proposed) and STLRRT extend KCBS with STL-guided specifications, while KcBOT and KRRT are KCBS-based methods implemented in C++ without STL monitoring. The priority-based approaches PPcBOT and PPRRT are also implemented in C++. In contrast, PBS+STGCS, SP+STGCS, and PP+STGCS are convex optimization–based planners implemented in Python and solved with the MOSEK optimizer, while SP+STRRT* represents a sampling-based STL variant developed in Python.

Experiments were conducted across four representative environments (Fig.~\ref{fig:bench}) with robot team sizes of 2, 4, 6, 8, 10, 12, and 50. Env.1 is an open workspace designed to evaluate baseline navigation efficiency. Env.2 introduces bottleneck corridors that require cooperative conflict resolution. Env.3 is a cluttered forest environment with dense obstacles and narrow passages, stressing scalability under constrained maneuvering. Finally, Env.4 is a canonical bugtrap environment combining wide-open and congested regions, requiring both global coordination and local adaptability.

\textbf{Success Rates.}
The proposed STLcBOT consistently achieved 100\% success across all environments and team sizes, demonstrating strong robustness to both structural complexity and increasing density. Among the baselines, KcBOT and PPcBOT also maintained high completeness. By contrast, RRT-based methods exhibited significant degradation as the number of robots increased. In Env.3, both KRRT and STLRRT failed beyond 12 robots, while in Env.2, KRRT degraded to about 80\% success and STLRRT dropped as low as 40\% at 50 robots. PPRRT proved comparatively more resilient, though its performance also declined in dense and cluttered scenarios. Convex optimization–based planners struggled severely in cluttered environments. In particular, all STGCS-based methods failed to find any feasible solutions in Env.~3, where narrow passages and high density created spatiotemporal bottlenecks. SP+STGCS and RP+STGCS broke down beyond 4 robots even in simpler environments. SP+STRRT* achieved limited success only in the empty map but collapsed in Env.4 after 2 robots. PBS+STGCS performed adequately in Env.1 and showed limited success in Env.4, but failed in more challenging maps, particularly Env.2.

\textbf{Runtime.}
STLcBOT and KcBOT achieved runtimes below $1$ second in Envs.~1 and 2 across all team sizes, with the sole exception of 50 robots in Env.~2, confirming their efficiency. Despite the additional STL monitoring overhead, STLcBOT remained tractable by pruning conflicts through robustness metrics. In contrast, RRT-based planners exhibited exponential growth with team size, exceeding $10^{1}$ seconds at 50 robots, while STLRRT and KRRT consistently occupied a middle ground between STLcBOT and the slower PPRRT. In Envs.2 and 3, cBOT-based planners such as STLcBOT and KcBOT clearly outperformed the RRT-based methods. STGCS-based planners were not competitive in runtime, as their reliance on large-scale convex optimization caused rapid escalation even in relatively simple environments. In Env.3, runtime analysis was moot for STGCS-based methods since none succeeded, whereas STLcBOT and KcBOT remained tractable. In Env.4, RRT-based methods achieved faster runtimes than cBOT-based ones, albeit with substantially longer trajectory length.

\textbf{Trajectory Efficiency.}
The proposed STLcBOT produced short, specification-compliant trajectories that scaled sub-linearly with team size and remained below 1200 m in Env.4, even with 50 robots. KcBOT and PPcBOT achieved similarly compact paths, whereas RRT-based methods exhibited severe path inflation, with lengths exceeding 3000m in Env.1 and 2500m in Env.2. In Env.3, the cBOT-based planners generated slightly longer trajectories than RRT-based methods at 50 robots, but the margin was negligible compared to their significantly higher success rates. Convex optimization–based methods occasionally produced competitive path lengths; however, their frequent failures in cluttered environments, particularly Env.3, prevented consistent evaluation.

These results highlight that the proposed STLcBOT framework delivers the most reliable balance between completeness, runtime efficiency, and trajectory quality. Unlike RRT-based planners, which degrade under density and structural complexity, and unlike STGCS-based methods, which fail in cluttered maps such as Env.3, STLcBOT ensures consistent, scalable, and specification-compliant multi-robot planning across all environments and large team sizes.

\section{Conclusion}
We presented a two-stage framework for multi-robot trajectory planning under Signal Temporal Logic (STL) specifications while respecting kinodynamic constraints. At the single-robot level, the proposed constrained Bayesian optimization–based tree search (cBOT) efficiently learns local cost maps and feasibility constraints to generate shorter, collision-free trajectories. At the multi-robot level, the STL-KCBS algorithm integrates STL monitoring into conflict-based search to ensure specification satisfaction with scalability and probabilistic completeness. We conducted extensive testing across diverse environments and team sizes, along with real-world experiments using ground and surface vehicles. Our planner generates shorter and smoother trajectories compared to existing RRT-based approaches. Moreover, our framework scales effectively to larger problems, successfully solving planning challenges for teams of up to 50 robots in complex environments where existing exact STL planners fail beyond 6 robots. 
% These experiments confirmed our approach's robustness, efficiency, and practical value for scalable multi-robot planning.
Future research directions include developing theoretical foundations for the proposed framework and expanding its application scope to encompass task and motion planning domains.

\begingroup
\footnotesize
\bibliographystyle{ieeetr}
\bibliography{ref}
\endgroup

\end{document}